\documentclass[runningheads]{llncs}
\usepackage[T1]{fontenc}
\usepackage{graphicx}
\usepackage{subcaption}
\usepackage{xcolor}
\usepackage{listings}
\usepackage{minted}
\usepackage{eso-pic}
\definecolor{lightgray}{gray}{0.5}

\AtBeginDocument{
  
}

\usepackage{xcolor}
\usepackage{listings}

\definecolor{goal}{RGB}{0,153,153}   
\definecolor{action}{RGB}{0,102,204}      
\definecolor{asComment}{RGB}{128,128,128}  

\begin{document}

\AddToShipoutPictureBG*{
  \AtPageUpperLeft{
    \raisebox{-1.5cm}{
      \makebox[\paperwidth]{
        \begin{minipage}{0.9\paperwidth}
          \centering\footnotesize
          \textcolor{lightgray}{This is an open-access, author-archived version of a manuscript published in European Conference on Multi-Agent Systems 2025.}
        \end{minipage}
      }
    }
  }
}
\AddToShipoutPictureBG*{
  \AtPageLowerLeft{
    \raisebox{1cm}{
      \makebox[\paperwidth]{
        \begin{minipage}{0.9\paperwidth}
          \centering\footnotesize
          \textcolor{lightgray}{©2025 Authors. This is the author’s version of the work. It is posted here for your personal use. Not for redistribution. \\
          The definitive Version of Record is published in the Proceedings of the European Conference on Multi-Agent Systems (EUMAS 2025).}
        \end{minipage}
      }
    }
  }
}
\title{Embedding Autonomous Agents in Resource-Constrained Robotic Platforms}
\author{Negar Halakou\inst{1}\orcidID{0009-0002-2783-9992} \and
Juan F. Gutierrez\inst{1}\orcidID{0000-0001-8509-8075} \and
Ye Sun\inst{1}\orcidID{0009-0009-4635-5652} \and
Han Jiang\inst{1}\orcidID{0009-0006-6517-7503} \and
Xueming Wu\inst{1}\orcidID{0009-0009-1220-5859} \and
Yilun Song\inst{1}\orcidID{0009-0005-3858-8676} \and
Andres Gomez\inst{1}\orcidID{0000-0002-5825-3567}}
\authorrunning{N.Halakou Author et al.}
\institute{Institut für Datentechnik und Kommunikationsnetze, TU Braunschweig, Germany
\email{\{negar.halakou, juan-felipe.gutierrez-gomez, ye.sun1, xueming.wu, h.jiang, yilun.song, andres.gomez\}@tu-braunschweig.de}\\
}
\maketitle              
\begin{abstract}
Many embedded devices operate under resource constraints and in dynamic environments, requiring local decision-making capabilities. 
Enabling devices to make independent decisions in such environments can improve the responsiveness of the system and reduce the dependence on constant external control. 
In this work, we integrate an autonomous agent, programmed using AgentSpeak, with a small two-wheeled robot that explores a maze using its own decision-making and sensor data. 
Experimental results show that the agent successfully solved the maze in 59 seconds using 287 reasoning cycles, with decision phases taking less than one millisecond. 
These results indicate that the reasoning process is efficient enough for real-time execution on resource-constrained hardware. 
This integration demonstrates how high-level agent-based control can be applied to resource-constrained embedded systems for autonomous operation. 

\end{abstract}

\section{Introduction}

Resource-constrained robotic platforms play an important role in enabling low-cost, large-scale deployments in many application scenarios.
Due to their limited size, memory, and processing power, these systems tend to use centralized processing. 
By only perceiving information locally and reasoning remotely, these systems suffer from limited autonomy and longer latencies in the actuation loop.

As the computational capacity of microcontrollers increases, researchers have begun introducing techniques like onboard machine learning, which can efficiently extract better information from its environment. 
In \cite{hao2025favbot}, Hao et al. propose a microrobot weighing less than 22 grams, including a camera, microcontroller, and actuator. 
The microrobot can detect a target and follow it, using only local data in a closed feedback loop. 
The entire logic, however, is implemented using ad-hoc C/C++ code and exhibits only reactive behavior.

Agent-oriented programming can bring many advantages to robotic platforms, facilitate the development of autonomous reasoning, swarm intelligence, among others. 
While existing Java-based frameworks have already integrated robotic control into the AgentSpeak language \cite{wesz2015integrating}, these are not compatible with resource-constrained robotic platforms based on microcontrollers (MCUs). 
In order to achieve this, specialized toolchains and frameworks have been developed to cope with the limited memory and processing capabilities of MCUs \cite{vachtsevanou2023embedding}. 

In this demo paper, we show the feasibility of embedding autonomous BDI agents in a microcontroller-based two-wheeled robotic platform and tasking the agent to solve a line-following maze. In order to achieve this, we have defined and implemented a minimal API for controlling our robot's movement. This hardware-dependent code can be easily reused by any AgentSpeak script code running on our robot. 
Lastly, we have experimentally validated our autonomous agent framework, showing that our reasoning cycles can execute fast enough for our robot to efficiently solve mazes.

\section{System Design}
An autonomous agent is an intelligent system capable of perceiving its environment through sensors and acting on that environment using actuators~\cite{bordini2007programming}. 
Unlike the imperative programming typically used in embedded systems, agents operate autonomously, making rational decisions, based on their perceptions and internal knowledge to achieve their goals. 
What sets agents apart is their autonomy. 
They do not require constant direction from humans or other systems. 

The Belief-Desire-Intention (BDI) model is one approach to implement the autonomous and intelligent behavior of agents. 
It provides a structured approach for designing autonomous agents by defining their behavior through three key elements: beliefs, desires, and intentions. 
Autonomous agents can be programmed using the BDI paradigm as implemented in the AgentSpeak language and executed by the Jason interpreter.
A simplified variant of Jason is used as the basis for the Embedded-BDI framework \cite{santos2022programacao}. 
This framework consists of a translation engine that converts AgentSpeak programs into optimized C++ code, a runtime library responsible for executing the agent’s reasoning process, and hardware-dependent code; 
together, these components form an executable binary. 
The workflow of the framework involves programming the agent’s deliberation logic in AgentSpeak, while the perception and action functions, along with other hardware-specific code, are implemented in C/C++ \cite{william2022increasing}. 

Building on these principles, in this work, we embed a BDI agent into a robotic platform to enable autonomous maze exploration. 
The agent forms beliefs based on its perception of the environment through line sensors, which detect intersections and path availability. 
Its main desire is to reach the goal, while intentions are generated and dynamically updated as navigation plans following the left-hand rule. 
The BDI agent manages navigation and decision-making by reasoning over its beliefs and selecting appropriate actions. 
These intentions are executed by the robot’s actuators, enabling it to move forward or turn as needed. 

\section{System Implementation}
We utilize a Pololu 3pi+ 2040 robot (Standard Edition), which features an RP2040 microcontroller with 264\,kB of onchip SRAM and 16\,MB of external flash. 
The robot includes five downward-facing reflectance sensors for line following, and two DC micro metal gear motors that independently drive the left and right wheels, enabling precise movement control via PWM signals.
To implement autonomous decision-making, we integrated the Embedded-BDI framework\footnote{https://embedded-bdi.github.io/} with Pololu's bare-metal firmware. 
This integration enables us to define the agent logic using AgentSpeak. 
The corresponding code is shown in Listing~\ref{asl_code}. \\ 

\begin{lstlisting}[caption={BDI agent logic in AgentSpeak for left-hand rule navigation.}, label={asl_code}]
!solve_maze.

+!solve_maze : at_intersection <- 
    !!handle_intersection.

+!solve_maze <- 
    follow_segment; 
    !!solve_maze.

+!handle_intersection <- 
    check_situation; 
    !!make_decision;
    !!solve_maze.

+!make_decision : goal_found <-stop.
+!make_decision : path_left <- turn_left. 
+!make_decision : path_straight <- forward. 
+!make_decision : path_right <- turn_right. 
+!make_decision <- rotate_180.

\end{lstlisting}

\subsection{BDI-Based Maze Solving Behavior}
After powering on, the robot initializes its hardware components. 
The process begins when the user presses button A, which triggers sensor calibration to adjust the line sensors for accurate path detection. 

Once the system is ready, the BDI agent starts execution with the initial goal \texttt{solve\_maze}. 
The robot’s behavior is managed in a cyclic manner through the agent’s reasoning cycle. 
In each step, the agent evaluates whether the robot is currently at an intersection. 
When an intersection is detected, the agent posts a new goal \texttt{handle\_intersection} using the \texttt{!!} operator. 
Although this creates a separate intention, it typically begins execution immediately within the same cycle. 
During this intention, the agent executes \texttt{check\_situation} to interpret sensor data and form beliefs such as \texttt{goal\_found}, \texttt{path\_left}, or \texttt{path\_right}. 
Based on these beliefs, the agent proceeds with \texttt{make\_decision}, selecting the appropriate action according to the left-hand rule. 

Based on this rule, when the robot arrives at an intersection, it first checks whether a line is detected on the left. 
If so, it turns left. 
If no line is detected on the left, it attempts to move forward. 
If no visible path is available ahead, it then checks the right side. 
If no path is detected in any direction, the robot performs a 180-degree turn to continue exploration. 
If no intersection is detected, the agent continues along the current path by executing \texttt{follow\_segment}. 
This loop of perception, reasoning, and action repeats until the goal is found, at which point the robot stops and the task is considered successfully completed. 

\begin{figure}[!h]
    \centering
    \begin{subfigure}[b]{0.38\linewidth}
        \centering
        \includegraphics[width=0.85\textwidth]{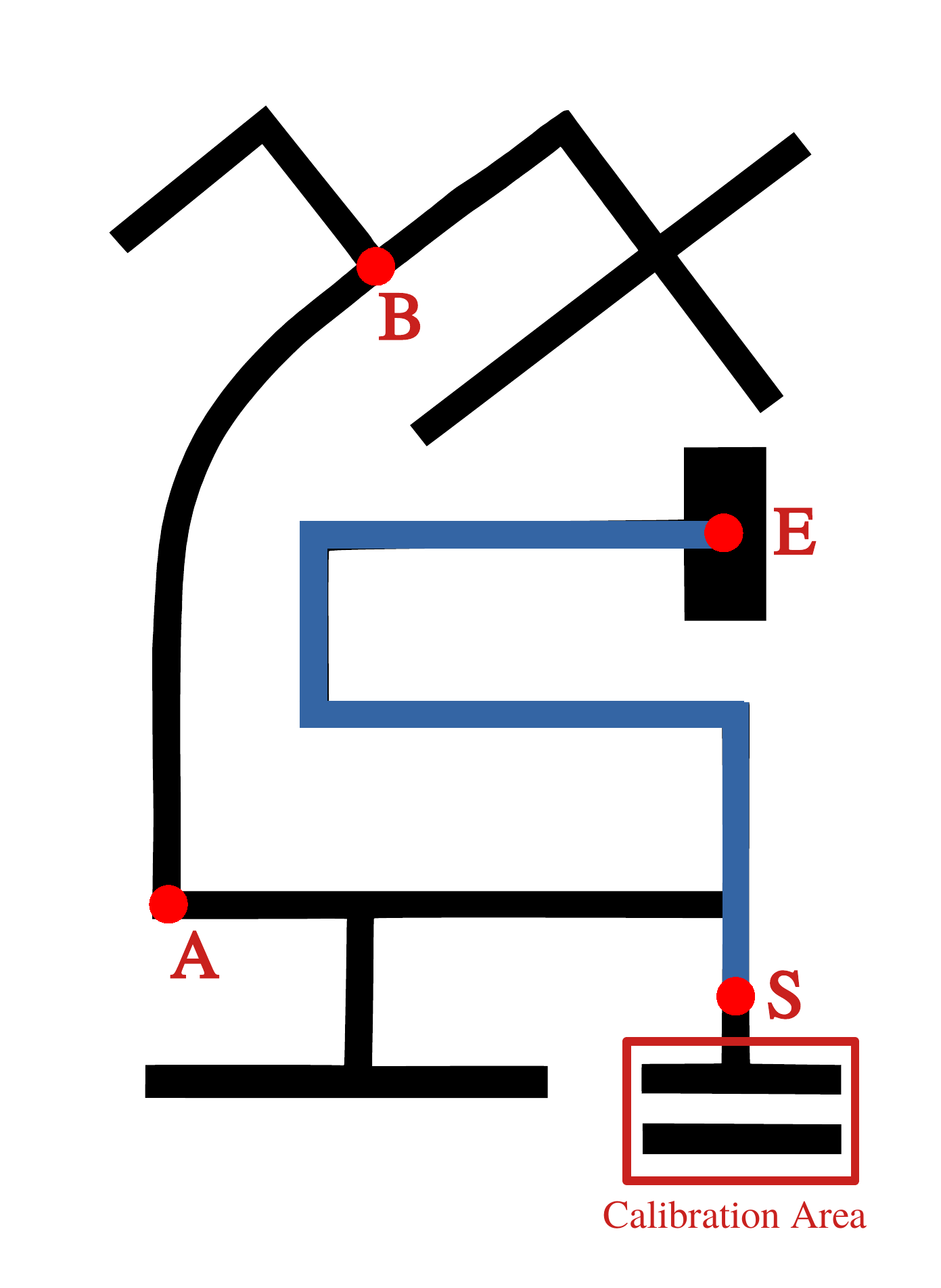} 
            \caption{Digital design of the maze.}
        \label{fig:maze_design_sub}
    \end{subfigure}
    \hfill
    \begin{subfigure}[b]{0.38\linewidth}
        \centering
        \includegraphics[width=0.8\textwidth]{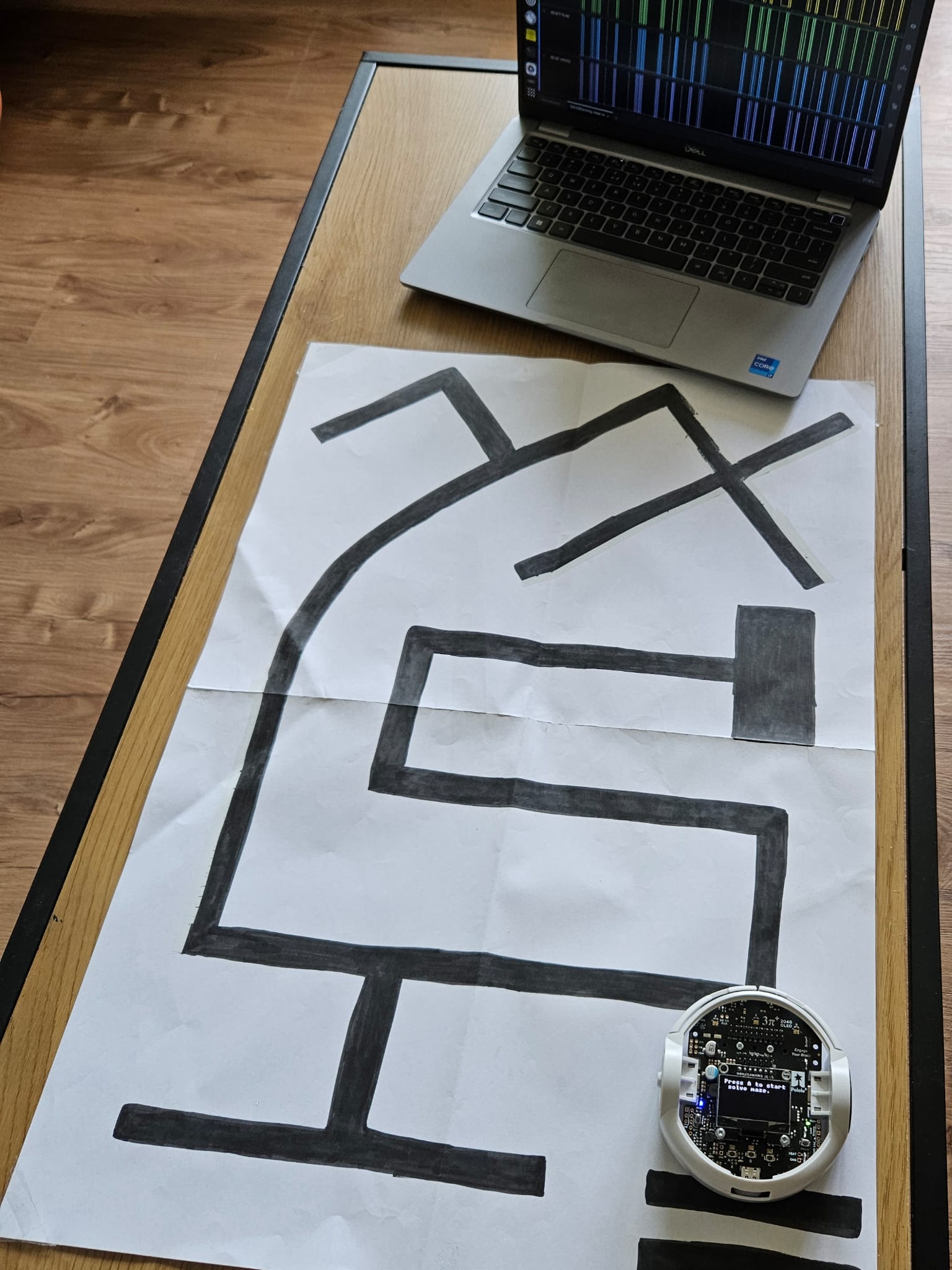} 
        \caption{Real-world implementation. }
        \label{fig:real_maze_sub}
    \end{subfigure}
    \caption{The designed maze used for planning (left) and the physical implementation with the robot (right).}
    \label{fig:maze_design}
\end{figure}

\vspace{-2em}
\section{Demonstration}

The effectiveness of the AgentSpeak-based algorithm was demonstrated using the maze shown in Figure~\ref{fig:maze_design}. 
The maze was specifically designed to first calibrate the line sensors at the starting area, enabling the robot to distinguish between black and white surfaces. 
Its layout forces the robot to follow the longest possible path before reaching the goal. 
A demonstration video and AgentSpeak-based maze solver robot software are available in a public repository\footnote{Demo-Paper Repo: \url{https://git.rz.tu-bs.de/ida/rosy/public/publication-repos/eumas-2025-mas-pololu-demo-paper}}. 
To analyze the agent’s behavior during maze solving, we measured the execution times of three main components in its reasoning cycle: belief update, plan selection, and intention execution. 

\begin{figure}[t]
    \centering
    \includegraphics[width=\linewidth]{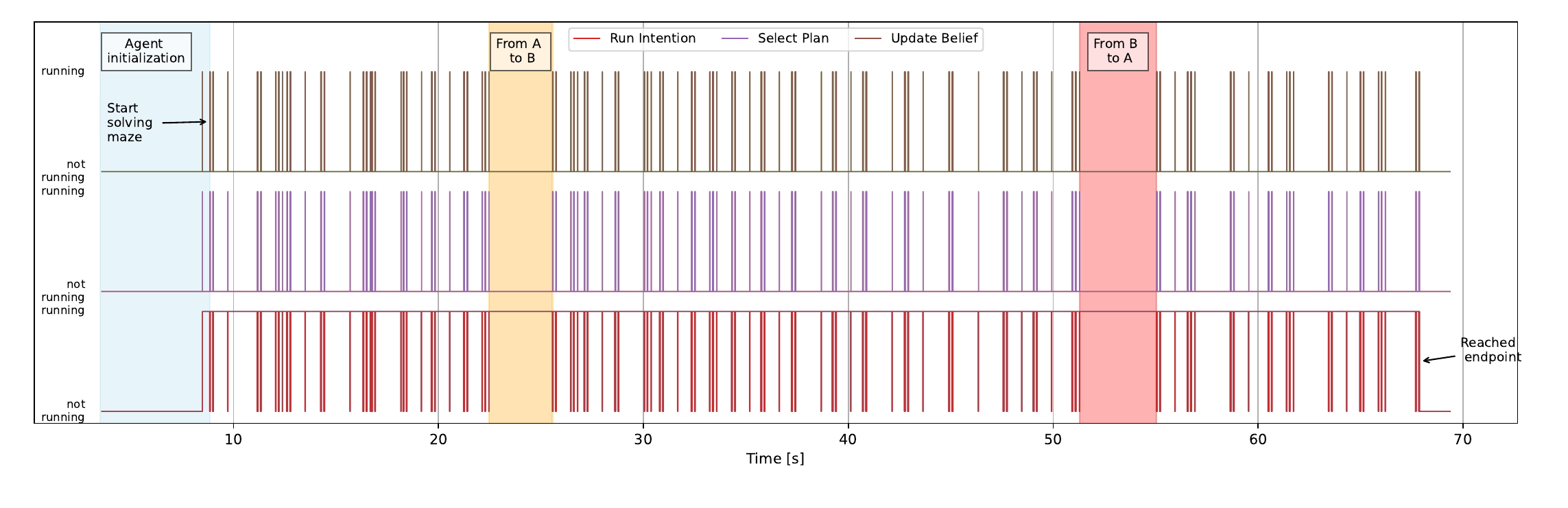}
    \caption{Annotated GPIO activity during the execution cycle of an agent-based maze-solving robot.}
    \label{fig:gpio_signals}
\end{figure}

These measurements were recorded as digital signals, with 1 indicating the function is running and 0 indicating it is not, as shown in Figure~\ref{fig:gpio_signals}.
Based on these measurements, the total time taken to solve the maze was approximately 59\,s. 
This duration corresponds to the trajectory from point~S (start point) to point~E (endpoint) along the longest path, during which the reasoning cycle was executed 287 times. 

On average, the belief update phase took 0.004\,ms (with a maximum of 0.022\,ms), the plan selection phase took 0.024\,ms (with a maximum of 0.153\,ms), and the intention execution phase lasted 197\,ms on average (with a maximum of 3744\,ms). 
The relatively long duration of the intention execution phase is mainly due to sensor readings and motor control operations, which are influenced by the specific structure of the maze. 
For instance, when the robot traveled from point A to point B, this phase lasted about 3.1\,s, while the return from B to A took approximately 3.7\,s. 
During both segments, the same plan was executed continuously, as no intersections were encountered. 
In contrast, the belief update and plan selection phases were consistently short. 
These short execution times suggest that the reasoning cycle is efficient enough for real-time decision-making, even at higher robot speeds, without adversely affecting task performance. 
We also evaluated memory usage. The compiled binary, which includes the translated AgentSpeak code, the runtime and the hardware-specific code, used only 5.44\% of the Flash memory and 6.25\% of the RAM.

\section{Conclusions}

In this demonstration, we embedded a BDI agent on a two-wheeled robot using a reusable API within the Embedded-BDI platform. 
Experiments showed that action execution dominated cycle time, while decision-making was computationally efficient. 
This confirms the feasibility of real-time autonomy on constrained hardware. 
As future work, the robot will communicate its discovered path, using AgentSpeak, to another AgentSpeak-based robot, enabling the second robot to traverse the shortest route to the goal. 

\bibliographystyle{splncs04}
\bibliography{refs}

\end{document}